\documentclass[prologue,table,sigconf]{acmart}
\AtBeginDocument{%
  }

\usepackage[capitalize]{cleveref}
\crefname{section}{Sec.}{Secs.}
\Crefname{section}{Section}{Sections}
\Crefname{table}{Table}{Tables}
\crefname{table}{Tab.}{Tabs.}

\usepackage[normalem]{ulem}
\useunder{\uline}{\ul}{}
\usepackage{multirow}
\def\model{PP-Motion}
\usepackage[skip=7pt]{caption}

\setcopyright{acmlicensed}              % 显示版权

\copyrightyear{2025}
\acmYear{2025}
\setcopyright{acmlicensed}\acmConference[MM '25]{Proceedings of the 33rd
ACM International Conference on Multimedia}{October 27--31, 2025}{Dublin,
Ireland}
\acmBooktitle{Proceedings of the 33rd ACM International Conference on
Multimedia (MM '25), October 27--31, 2025, Dublin, Ireland}
\acmDOI{10.1145/3746027.3754940}
\acmISBN{979-8-4007-2035-2/2025/10}

\acmSubmissionID{1224}

\begin{document}

\title[PP-Motion: Physical-Perceptual Fidelity Evaluation for Human Motion Generation]{PP-Motion: Physical-Perceptual Fidelity Evaluation \\for Human Motion Generation}

\author{Sihan Zhao}
\authornote{Equal contribution.}
\affiliation{%
  \institution{Tsinghua University}
  \city{Beijing}
  \country{China}
}
\orcid{0009-0002-0824-6358}
\email{zhaosh024@gmail.com}

\author{Zixuan Wang}
\authornotemark[1]
\affiliation{%
  \institution{Tsinghua University}
  \city{Beijing}
  \country{China}
}
\orcid{0000-0001-7291-6198}
\email{wangzixu21@mails.tsinghua.edu.cn}

\author{Tianyu Luan}
\affiliation{%
  \institution{University at Buffalo}
  \city{Buffalo}
  \state{NY}
  \country{USA}
}
\orcid{0000-0001-7333-1052}
\email{tianyulu@buffalo.edu}
\authornote{Co-corresponding authors.}

\author{Jia Jia}
\affiliation{%
  % \department{BNRist}
  \institution{BNRist, Tsinghua University}
  % \department{Key Laboratory of Pervasive Computing}
  \institution{Key Laboratory of Pervasive Computing, Ministry of Education}
  \city{Beijing}
  \country{China}
}
\orcid{0009-0005-8449-278X}
\email{jjia@tsinghua.edu.cn}
\authornotemark[2]

\author{Wentao Zhu}
\affiliation{%
  \institution{Eastern Institute of Technology, Ningbo}
  \city{Ningbo}
  \country{China}
}
\orcid{0000-0002-5483-0259}
\email{wtzhu@eitech.edu.cn}

\author{Jiebo Luo}
\affiliation{%
  \institution{University of Rochester}
  \city{Rochester}
  \state{NY}
  \country{USA}
}
\orcid{0000-0002-4516-9729}
\email{jluo@cs.rochester.edu}

\author{Junsong Yuan}
\affiliation{%
  \institution{University at Buffalo}
  \city{Buffalo}
  \state{NY}
  \country{USA}
}
\orcid{0000-0002-7901-8793}
\email{jsyuan@buffalo.edu}

\author{Nan Xi}
\affiliation{%
  \institution{University at Buffalo}
  \city{Buffalo}
  \state{NY}
  \country{USA}
}
\orcid{0000-0002-7334-7772}
\email{nanxi@buffalo.edu}

\setlength{\floatsep}{8pt}      % 两个float之间（如两个table/figure）
\setlength{\textfloatsep}{10pt} % float与正文之间
\begin{abstract}
% Human pose generation has found widespread applications in AR/VR, film, sports, and medical rehabilitation, offering a cost-effective alternative to traditional motion capture systems. Evaluating the fidelity of these generated pose sequences is a crucial task. Although previous approaches have advanced pose sequence fidelity evaluation using human perception, there remains an inherent gap between human-perceived fidelity and physical feasibility. Moreover, the subjective and coarse binary labeling further undermines the development of a robust data-driven metric. To solve these problems, we propose a novel evaluation method that measures pose fidelity by computing the minimum adjustments required for a pose to comply with physical laws. This approach produces fine-grained, continuous physical alignment annotations that serve as objective ground truth. Our framework integrates these annotations with coarse human perceptual labels and employs correlation-based loss functions, including Pearson's correlation loss and Spearman's ranking order loss, to effectively capture underlying physical priors. Experimental results demonstrate that our metric not only adheres to physical laws but also aligns well with human perception of pose sequence fidelity. 
%wzx%
Human motion generation has found widespread applications in AR/VR, film, sports, and medical rehabilitation, offering a cost-effective alternative to traditional motion capture systems. However, evaluating the fidelity of such generated motions is a crucial, multifaceted task. Although previous approaches have attempted at motion fidelity evaluation using human perception or physical constraints, there remains an inherent gap between human-perceived fidelity and physical feasibility. Moreover, the subjective and coarse binary labeling of human perception further undermines the development of a robust data-driven metric. We address these issues by introducing a physical labeling method. This method evaluates motion fidelity by calculating the minimum modifications needed for a motion to align with physical laws. With this approach, we are able to produce fine-grained, continuous physical alignment annotations that serve as objective ground truth. With these annotations, we propose PP-Motion, a novel data-driven metric to evaluate both physical and perceptual fidelity of human motion. To effectively capture underlying physical priors, we employ Pearson's correlation loss for the training of our metric. Additionally, by incorporating a human-based perceptual fidelity loss, our metric can capture fidelity that simultaneously considers both human perception and physical alignment. Experimental results demonstrate that our metric, PP-Motion, not only aligns with physical laws but also aligns better with human perception of motion fidelity than previous work. Code is available at: https://github.com/Sarah816/PP-Motion.
%wzx end%
%wzx 0411comment 感觉要点一下PP motion，所以改了一下%
\end{abstract}




%   \includegraphics[width\textbf{}=\textwidth]{sampleteaser}

\begin{teaserfigure}
  \centering
  \includegraphics[width=0.86\linewidth]{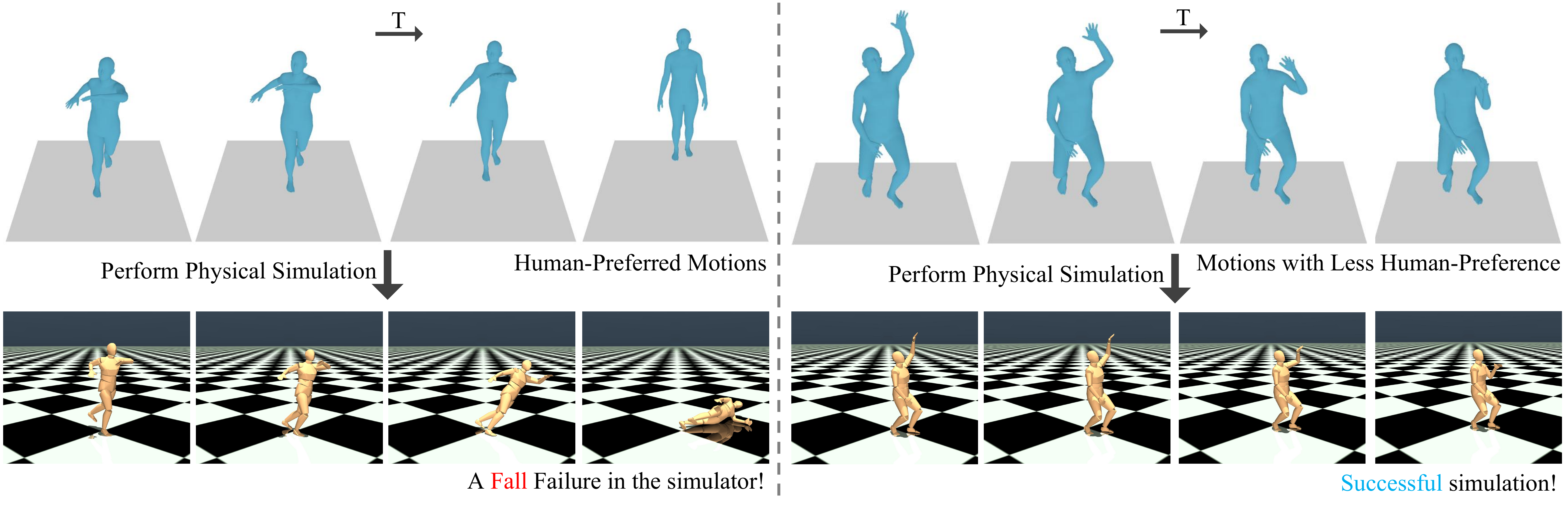}
  \caption{A motion that looks realistic does not necessarily mean it is physically feasible. Top-left: Motion appears realistic and semantically meaningful to the human eye, yet fails in physics simulation, resulting in a fall (bottom-left). Top-right: Unnatural motion in human perception executes successfully in simulation (bottom-right). This reveals a discrepancy between human perception and physical laws.}
  \Description{A teaser figure to show our motivation.}
  \label{fig:teaser}
\end{teaserfigure}


\maketitle

\section{Introduction}
Human motion generation has found widespread applications in modern industrial production. Whether in AR/VR games, film, content creation, sports, or medical rehabilitation, generated motions can replace complex motion capture systems and on-site filming with actors. Realistic human motion generation can produce large quantities of poses at very low cost, potentially saving significant labor and filming expenses. Therefore, evaluating the fidelity of generated motions is a problem closely tied to real-world applications. However, it remains a challenging task to design a comprehensive evaluation metric due to the multifaceted influence factors.

% Previous methods, such as \cite{motioncritic2025}, have made good progress in assessing pose sequence fidelity. That work introduces a dataset where human subjects judge the fidelity of a pose. These ratings are later used as labels to train a pose sequence fidelity metric that aligns with human perception. While human judgments of pose sequence fidelity are reasonably accurate and practically useful, the fundamental standard for pose fidelity should not be based solely on human perception. It is even more important to consider whether the pose conforms to physical laws. A pose that looks realistic does not necessarily mean it is physically feasible. For example, as shown in \cref{fig:teaser}, the pose sequence in the top-left appears realistic and semantically meaningful to human eyes, yet when simulated in a physics engine, the motion cannot be completed and results in a fall on the ground(bottom-left). Conversely, the pose sequence on the top-right may look unusual and meaningless, but it can be executed well in a physics simulation (bottom-right). These examples reveal a discrepancy between human perception and physical laws. Moreover, human annotations of fidelity are subjective; different annotators may have difficulty to consistently quantify the fidelity of the same pose. In \cite{motioncritic2025}, the dataset employs a binary “good-bad” classification to avoid the quantification issues of human labeling. However, such coarse labeling lacks fine-grained information and poses challenges for learning a data-driven metric.
%wzx%
Previous methods, such as MotionCritic~\cite{motioncritic2025}, have made good progress in assessing human motion fidelity. MotionCritic introduces a dataset, \textit{MotionPercept}, where human subjects judge the fidelity of a motion. These ratings are later used as labels to train a motion fidelity metric that aligns with human perception. Although human perception of motion fidelity is reasonably accurate and practically useful, the fundamental standard for motion fidelity should not be based solely on human perception. It is even more important to consider whether the motion conforms to physical laws. A motion that looks realistic does not necessarily mean it is physically feasible. For example, as shown in \cref{fig:teaser}, the motion on the top-left appears realistic and semantically meaningful to human eyes, yet when simulated in a physics engine, the motion cannot be completed and results in a fall on the ground (bottom-left). Conversely, the motion on the top-right may look unusual and meaningless, but it can be executed well in a physics simulation (bottom-right). These examples reveal a discrepancy between human perception and physical laws. Moreover, human annotations of fidelity are subjective; different annotators may have difficulty quantifying the fidelity of the same motion consistently. In MotionCritic~\cite{motioncritic2025}, the dataset employs a binary ``better/worse'' classification to avoid the quantification issues of human labeling. However, such coarse labeling lacks fine-grained information and poses challenges for learning a data-driven metric.
%wzx end%

To address these two issues, we propose a data-driven method to evaluate motion fidelity that aligns with physical laws. We achieve this by calculating the minimum distance between the test motion and a motion that complies with physical laws. A small minimal distance indicates a high fidelity of the input motion, whereas a large minimal distance implies low fidelity. With this physically grounded definition of fidelity, we can establish fine-grained continuous labels for physical law alignments. In addition, we design a framework that trains a metric, named PP-Motion, by simultaneously utilizing fine-grained physical labels and coarse, discrete human perceptual labels. By designing loss functions that better suit the fine-grained labels, we can more effectively learn the underlying physical law priors. 
% This approach not only allows our metric to align more closely with physical laws but, given the inherent correlation between human perception and physical feasibility, it also has the potential to achieve better alignment with human perception through learning physical principles.
With this approach, we are able to train our metric to better align with physical laws. Furthermore, since human perception is inherently correlated with physical feasibility, this approach also improves the potential for metric design to align with human judgments by learning physical principles.

Specifically, we adopt the motions generated in MotionCritic \cite{motioncritic2025} and create new physically aligned annotations for them. Inspired by PHC \cite{Luo2023PerpetualHC}, we refine every motion in our dataset using reinforcement learning with the help of a physics simulator.
We use this approach to make only minimal adjustments while making each motion conform to physical laws. We then compare the difference between the adjusted motion and the original motion to serve as the annotation for physical alignment in the dataset. Such annotations not only closely adhere to our definition of physical fidelity and offer strong interpretability, but also provide continuous, fine-grained labels. These fine-grained annotations offer rich information for the supervision of subsequent metric training. To better learn from these fine-grained physical annotations, we design loss functions based on data correlation, such as Pearson's correlation loss. Unlike previous classification losses, correlation loss can effectively capture the intrinsic correlations within the data rather than simply comparing categories or numeric values. Without the constraint on scales, the correlation loss can be more easily combined with existing human perception loss functions, enabling our metric to align with both human perception and physical laws, and providing the potential for mutual reinforcement between the two aspects.

In summary, our contributions are as follows:\vspace{-2mm}
\begin{itemize}
    \item We propose a novel fidelity evaluation method, \textbf{\model{}}, for human motions, which takes into account both physical feasibility and human perception. Our method can evaluate whether a motion is realistically aligned with physical laws and human perception.
    \item We define and design a fine-grained physical alignment annotation and provide this annotation for existing datasets. This annotation serves as fine-grained physical ground truth for training our metric and has the potential to benefit subsequent metric design.
    \item We design an effective learning framework that leverages these fine-grained physical annotations. By incorporating correlation-based loss functions (i.e., Pearson's correlation loss), our approach better learns the physical priors from the labels, while seamlessly combining with existing human perceptual loss functions. This design not only ensures that our metric adheres to physical laws but also has the potential to enhance human-percepted fidelity.
\end{itemize}

\begin{figure*}[ht]
  \centering
  \includegraphics[width=0.8\linewidth]{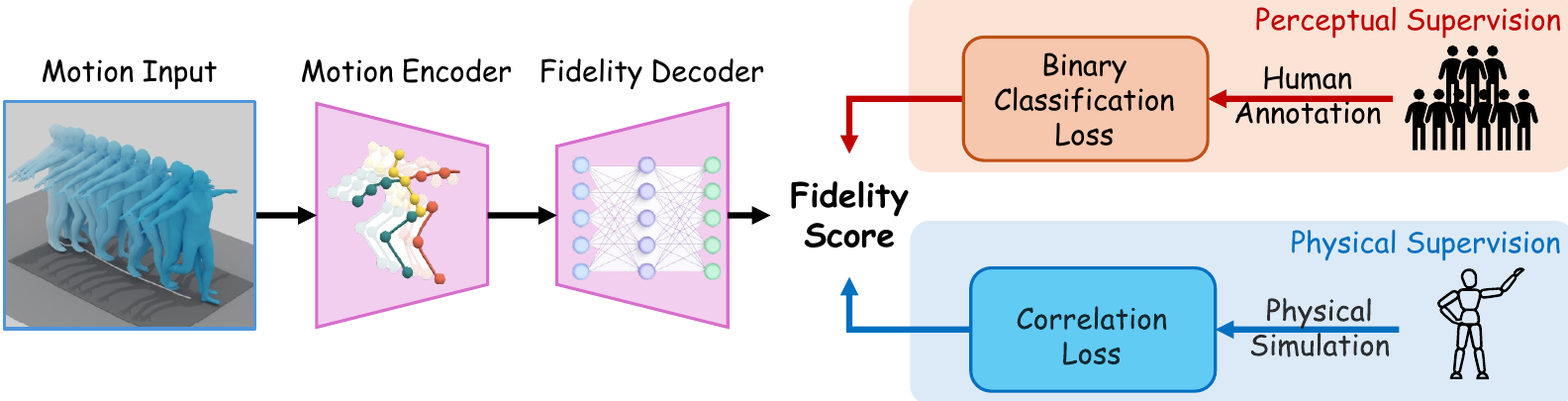}
  \caption{Our metric design and training pipeline. The network takes a human motion sequence as input, which is processed by a motion encoder to extract spatiotemporal features. These features are then decoded into a fidelity score by a fidelity decoder. The network is trained in a supervised fashion using fine-grained physical annotations alongside human perceptual labels. }
  \Description{Pipeline}
  \label{fig:pipeline}
\end{figure*}

\section{Related Work}

\subsection{Human Motion Generation} 
Human motion generation aims to automatically generate natural, fluent, and physically plausible human pose sequences, playing a central role in character animation, human-robot interaction, and embodied agents acting in complex environments. With wide applications and high practical value, motion generation is a foundational problem in both academic and industrial fields. Fueled by rapid advances in deep learning~\cite{lecun2015deep}, especially generative models~\cite{goodfellow2014generative,ho2020denoising,kingma2013auto,rezende2015variational,radford2018improving}, extensive research has focused on generating human motions from multimodal signals, including text, action, speech, and music. Action-to-motion~\cite{lucas2022posegpt,chen2023executing,TEACH:3DV:2022,guo2020action2motion} aims to synthesize motions from predefined action labels, evolving from retrieval-based to label-conditioned generative models. Text-to-motion~\cite{tevet2022human,tevet2022motionclip,petrovich2022temos,uchida2024mola,wang2024holistic,huang2024stablemofusion,sun2024lgtm,zhang2025motion,gao2024guess,pinyoanuntapong2024mmm,guo2024momask,zhang2024motiondiffuse,guo2022tm2t,petrovich22temos,zhai2023language} focuses on mapping natural language to motion, bridging linguistic semantics and physical embodiment. Beyond text, audio-driven motion generation has also seen progress. Music-to-dance methods~\cite{tseng2023edge,siyao2022bailando,tang2018dance,li2023finedance,li2021ai} synthesize motions aligned with musical features such as beat and style. Speech-to-gesture approaches~\cite{ginosar2019learning,yoon2020speech,habibie2021learning,bhattacharya2021speech2affectivegestures,qian2021speech,li2021audio2gestures,ao2022rhythmic,kucherenko2019analyzing} emphasize temporal alignment and semantic expressiveness to convey emotion. Further studies on controllable and editable motion generation~\cite{motionlcm,barquero2024seamless,hoang2024motionmix,zhang2023finemogen,wan2024tlcontrol,karunratanakul2024optimizing,xie2024omnicontrol,shafir2024human} focus on generating high quality long-term motions while maintaining maximum faithfulness to various multimodal control conditions. Considering the growing range of real-world applications and generative methods of motion generation, it’s critical to establish comprehensive evaluations for generated motions.

\subsection{Human Motion Evaluation}
Designing evaluation metrics for human motion is a complicated and challenging problem. Existing evaluation metrics can mainly be divided into three categories: (1) distance-based metrics, (2) human-perception-based metrics, and (3) physical plausibility metrics. The most commonly used evaluation metrics in early works are distance-based. Distance-based metrics such as Position, Velocity, and Acceleration Errors~\cite{kucherenko2019analyzing,wang2022humanise,ahuja2019language2pose,petrovich2022temos,athanasiou2022teach,kim2023flame,zhou2023ude,tang2018dance,ginosar2019learning,li2021audio2gestures,corona2020context,cao2020long,wang2021synthesizing,mao2022contact,huang2023diffusion,luan2021pc,luan2023high,luan2025scalable,luan2024divide,zhang2021learning} compare generated motions with ground truth, but struggle to capture the diversity of plausible motions. To address this, 
feature-based metrics~\cite{ghosh2021synthesis,zhou2023ude,yoon2020speech,habibie2021learning,bhattacharya2021speech2affectivegestures,qian2021speech,ao2022rhythmic,ghosh2023imos,gopalakrishnan2019neural} are proposed to provide more refined semantic abstraction to calculate distance for similarity assessment between generated and ground truth motions. Among all these distance-based metrics, Fréchet Inception Distance (FID) and average Euclidean distance (DIV) in feature space are widely used to evaluate motion quality and diversity~\cite{zhou2023ude,siyao2022bailando,li2021ai,li2023finedance,li2024lodge,guo2022generating,zhu2023human,jiang2023motiongpt,zhang2023generating,tevet2022human,chen2023executing,tseng2023edge,zhang2025motion}. However, FID and DIV measure the distance between or within distributions, making them unsuitable for evaluating the quality of a single action. Feature-space metrics also depend heavily on the effectiveness of the feature extractor, where certain features may lack interpretability. To better assess motion naturalness, smoothness, and plausibility, \citet{motioncritic2025} propose MotionCritic, a data-driven model trained on human-annotated motion preferences. Although MotionCritic aligns well with human perception, there remains a huge gap between motions deemed feasible by humans and those grounded in physical laws. For physical plausibility evaluation, researchers have proposed rule-based metrics such as foot-ground penetration~\cite{rempe2020contact,rempe2021humor,hassan2021stochastic,taheri2022goal}, foot contact~\cite{rempe2020contact,rempe2021humor,tseng2023edge}, foot skating rate~\cite{louis2025measuring,wu2022saga,araujo2023circle}, and floating~\cite{han2024reindiffuse}. However, these metrics are often heuristic, threshold-sensitive, and too limited to capture overall physical fidelity. Physdiff~\cite{yuan2023physdiff} proposes an overall physics error metric, but remains a naive aggregation of penetration, foot skating, and floating metrics.  \citet{li2024morph} further propose the imitation failure rate (IFR), using a physics engine to test whether a motion can be successfully simulated. However, IFR only gives a binary judgment, offering no graded assessment. Thus, a comprehensive evaluation method that jointly considers human perception and physical feasibility remains an open challenge.

\section{Method}
\subsection{Problem Formulation}
In our approach to designing a motion evaluation metric, the measurement system is defined as follows. For any human motion $x$, the aim is to establish a function $F(\cdot)$ that evaluates the fidelity $\hat{s}$ of the motion sequence:
\begin{equation}
    \hat{s} = F(x;\theta),
\end{equation}
%where $F(x;\theta)$ is a neural network architecture with $\theta$ as its parameters.
%wzx
where $F(x;\theta)$ is a neural network architecture with parameters $\theta$.

The subsequent sections introduce our methodology for both architecting the measurement function $F(x;\theta)$ and optimizing its parameters. The training procedure can be summarized by the optimization objective:
\begin{equation}
    \min_\theta{\mathcal{L_\text{prec}}(F(x;\theta), y_{\text{prec}})} + \lambda\mathcal{L_\text{phy}}(F(x;\theta), y_{\text{phy}}),
\end{equation}
where $\mathcal{L_\text{prec}}$ and $\mathcal{L_\text{phy}}$ are the perceptual and physics losses, $y_{\text{prec}}$ and $y_{\text{phy}}$ are the perceptual and physics supervision, and $\lambda$ is a balance weight for these two terms of loss functions. The objective is to jointly optimize the measurement accuracy for both physical and perceptual fidelity.

\subsection{Physical-Perceptual Motion Metric}
The design of our metric is illustrated in \cref{fig:pipeline}. Our network takes a human motion sequence as input. First, the motion is fed into a motion encoder to extract spatio-temporal features. A fidelity decoder is then used to decode these features into a fidelity score. We use annotations from two different sources, physical and perceptual, to supervise the metric training. On the physical side, we analyze motion fidelity using a physics simulator and generate fine-grained annotations for training supervision. A correlation loss between the metric output and the physical annotations encourages the metric to learn from those physical annotations effectively. Meanwhile, the network also learns fidelity from human annotations, ensuring that the fidelity score aligns closely with both physical annotation and human perception.

\textbf{Motion encoder.}
The motion encoder plays a critical role in determining the quality of the motion features. To extract both spatial and temporal information necessary for fidelity evaluation, we adopt a state-of-the-art spatio-temporal motion encoder. In our experiments, we follow the design proposed in \cite{zhu2023learning}. This encoder is built from $N$ dual-stream fusion modules, each containing branches for spatial and temporal self-attention and MLP. The spatial layers capture correlations among different joints within the same time step, while the temporal layers focus on the dynamics of individual joints. This dual-stream design effectively captures comprehensive features required for assessing motion fidelity.

\textbf{Fidelity decoder.}
We introduce a fidelity decoder module to interpret motion fidelity from the extracted features. Since our fidelity score is a fine-grained continuous value rather than a coarse classification, we design a fidelity decoder to extract the fidelity score from the features. Moreover, given the strong representation capability of our backbone, we can leverage the encoder to extract and fit the fidelity features during training, which allows us to simplify the fidelity decoder design. In practice, we adopt an MLP-based design for the fine-grained fidelity decoder. This simple design not only meets the requirement for producing detailed scores but also avoids imposing a significant extra burden on network training.

\textbf{Physical supervision.}
\label{sec:metric}
Physical supervision is a core module of our method, and it consists of two parts: physical accuracy labels and the corresponding supervision strategy.

To evaluate physical fidelity, we develop a scoring system based on feedback from a physics simulator. For a given input motion, we generate a motion that is as close as possible to the input while satisfying the physical constraints of the simulator (i.e., the nearest regularized motion). The difference between the input pose and the nearest regularized motion represents the motion’s physical rationality. For more details on generating the nearest regularized motion, please refer to \cref{sec:dataset}. Intuitively, if a pose that is initially physically implausible can be made physically reasonable with only minor adjustments, it is considered to have high physical fidelity (i.e., a small fidelity error). However, if major modifications are needed, the pose is considered to have low physical fidelity (i.e., a large fidelity error). This design enables fine-grained continuous annotation and measurement of physical fidelity.

Our approach to physical supervision primarily focuses on aligning the network’s output with the physical annotations. We have observed that, for designing a physically aligned fidelity score, the absolute value of the output score is less critical than its correlation with the physical labels. Therefore, our supervision targets the correlation between the fidelity score and the physical annotations rather than direct numerical differences as in traditional regression tasks. To this end, we use Pearson’s correlation loss as the core training loss, as detailed in \cref{sec:loss}.

\textbf{Perceptual supervision.}
% For human-aligned fidelity evaluation, our network also leverages the annotations and training strategy provided by \cite{motioncritic2025}. Specifically, the fidelity annotations used in this branch are derived from human subjects. During training, we use the binary ``better/worse" labels provided by these subjects to train a network using a classification loss, which outputs a corresponding ``better/worse" score for the input motion. Notably, even though we did not modify the original training strategy or annotations for this part, our joint training under both physical and perceptual supervision results in a metric that aligns with human perception even better than a model optimized solely for human alignment. This outcome demonstrates that physical and perceptual annotations are well aligned and that fine-grained physical alignment can further boost human perceptual alignment.
%wzx comment 这个classification network可以直接说吗，还是说是一个打分的network，force 好的比坏的打分高
%tianyu: 感觉你说的对，你直接改吧。update: 改了，看看改的对不对。
%想改成下面这样
%During training, given a pair of motions with "better" and "worse" labels provided by human subjects, we train a network using a perceptual loss, which encourages the model to assign higher scores to the "better" motion than to the "worse" motion. 
%wzx改
For human-aligned fidelity evaluation, our network also leverages the annotations and training strategy provided by \cite{motioncritic2025}. Specifically, the fidelity annotations used in this branch are derived from human subjects. During training, given a pair of motions with "better" and "worse" labels provided by human subjects, we train a network using a perceptual loss, which encourages the model to assign higher scores to the "better" motion than to the "worse" motion. Notably, even though we did not modify the original training strategy or annotations for this part, our joint training under both physical and perceptual supervision results in a metric that aligns with human perception even better than a model optimized solely for human alignment. This outcome demonstrates that physical and perceptual annotations are well aligned and that fine-grained physical alignment can further boost human perceptual alignment.
\subsection{Training Loss}
\label{sec:loss}
Our loss design is mainly divided into two parts: a perceptual loss based on binary human fidelity scoring labels (i.e., better/worse in each pair), and a physical loss based on continuous physical labels.

\textbf{Perceptual Loss.}
Our perceptual loss follows the design in \cite{motioncritic2025}. For a better-worse motion pair $x^{(h)}$ and $x^{(l)}$, the perceptual loss is defined as:
\begin{equation}
\mathcal{L}_{\text{percept}} = -\mathbb{E}_{(x^{(h)}, x^{(l)})} \left[\log \sigma\Bigl(F\bigl(x^{(h)}\bigr) - F\bigl(x^{(l)}\bigr)\Bigr)\right],
\end{equation}
where $F(\cdot)$ is our designed metric, and $\sigma(\cdot)$ is the sigmoid function.

\textbf{Physical Loss.}
Our physical loss is designed to learn from the fine-grained physical annotation. We use a Pearson's correlation loss~\cite{pearson1920plcc} to learn for physical annotation. The correlation loss is defined as:
\begin{equation}
\mathcal{L}_{\text{corr}}=-\frac{\sum_{i=1}^n(\hat{x}_i-\bar{\hat{x}})(x_i-\bar{x})}{\sqrt{\sum_{i=1}^n(\hat{x}_i-\bar{\hat{x}})^2} \sqrt{\sum_{i=1}^n(x_i-\bar{x})^2}},
\label{eq:plcc}
\end{equation}
where $\hat{x}_i$ and $x_i$ indicate the predicted and ground truth motion fidelity scores of the sample $i$ in the dataset. $n$ is the total number of samples in the dataset. The average predicted motion fidelity and ground truth motion fidelity are then defined as $\bar{\hat{s}}=\frac{1}{n}\sum_{i=1}^n\hat{s}_i$ and $\bar{s}=\frac{1}{n}\sum_{i=1}^n s_i$.

\textbf{Total Loss.} Our total loss function can then be represented as:
\begin{equation}
\mathcal{L}=\mathcal{L}_{\text{percept}} + \lambda\mathcal{L}_{\text{corr}},
\end{equation}
where $\lambda$ is the loss weight. The implementation details of model training are provided in the supplementary material.

\begin{table}
\caption{Annotation statistics on MotionPercept dataset.}
\centering
\resizebox{0.88\linewidth}{!}{%
    %\begin{tabular}{|c|c|c|c|c|c|c|c|c|c|c|c|c|c|}
    \begin{tabular}{l|c|c}
    \hline
    Annoation & MotionCritic & Ours \\
    \hline
    Grain & Binary & Continous \\
    Annoation type & Perceptual & Physical \\
    Categorization & Quadruples & Per-prompt\\
    Score Distribution & $p(0)=0.75, p(1)=0.25$ & $\sim N(0,1)$ \\ 
    \hline
    \end{tabular}%
}
\label{tab:dataset}
\end{table}

\begin{table}
\centering
    \caption{Quantitative results on imitating motion sequences of MotionPercept, which has three subsets: MDM-Train, MDM-Val, and FLAME. We use pose-based metrics to compare the imitation performance between using only the pretrained model and applying per-sequence fine-tuning. Recon. Err., MPJPE, and PA-MPJPE are measured in millimeters (mm).}
    \label{tab:ft_mdm}
    \resizebox{\linewidth}{!}{%
    \begin{tabular}{l|ccccc}
    \hline
    \rowcolor{gray!20}\multicolumn{6}{c}{MotionPercept-MDM-Train} \\
    \hline
    % \cline{2-6}
     & Recon. Err.$\downarrow$& MPJPE$\downarrow$& PA-MPJPE$\downarrow$& $e_{\text{acc}}$$\downarrow$ & $e_{\text{vel}}$$\downarrow$ \\
     \hline
    Whole dataset pretrain & 55.72 & 36.76 & 30.60 & 4.37 & 6.23  \\
    Per data fine-tune & \textbf{49.65} & \textbf{32.95} & \textbf{27.37} & \textbf{4.03} & \textbf{5.76} \\
    \hline
    \hline
    \rowcolor{gray!20}\multicolumn{6}{c}{MotionPercept-MDM-Val} \\ 
    \hline
     & Recon. Err.$\downarrow$& MPJPE$\downarrow$& PA-MPJPE$\downarrow$& $e_{\text{acc}}$$\downarrow$ & $e_{\text{vel}}$$\downarrow$ \\
     \hline
    Whole dataset pretrain & 55.49 & 36.88 & 30.68 & 4.27 & 6.07 \\
    Per data fine-tune & \textbf{50.90} & \textbf{34.10} & \textbf{28.20} & \textbf{4.16} & \textbf{5.89} \\
    \hline
    \hline
    \rowcolor{gray!20}\multicolumn{6}{c}{MotionPercept-FLAME} \\
    \hline
     & Recon. Err.$\downarrow$& MPJPE$\downarrow$& PA-MPJPE$\downarrow$& $e_{\text{acc}}$$\downarrow$ & $e_{\text{vel}}$$\downarrow$ \\
     \hline
    Whole dataset pretrain & 69.20 & 49.80 & 37.91 & 5.63 & 7.94 \\
    Per data fine-tune & \textbf{53.76} & \textbf{38.33} & \textbf{31.35} & \textbf{5.32} & \textbf{7.26} \\
    \hline
    \end{tabular}
    }
\end{table}

\begin{figure*}[ht]
  \centering
  \includegraphics[width=0.9\linewidth]{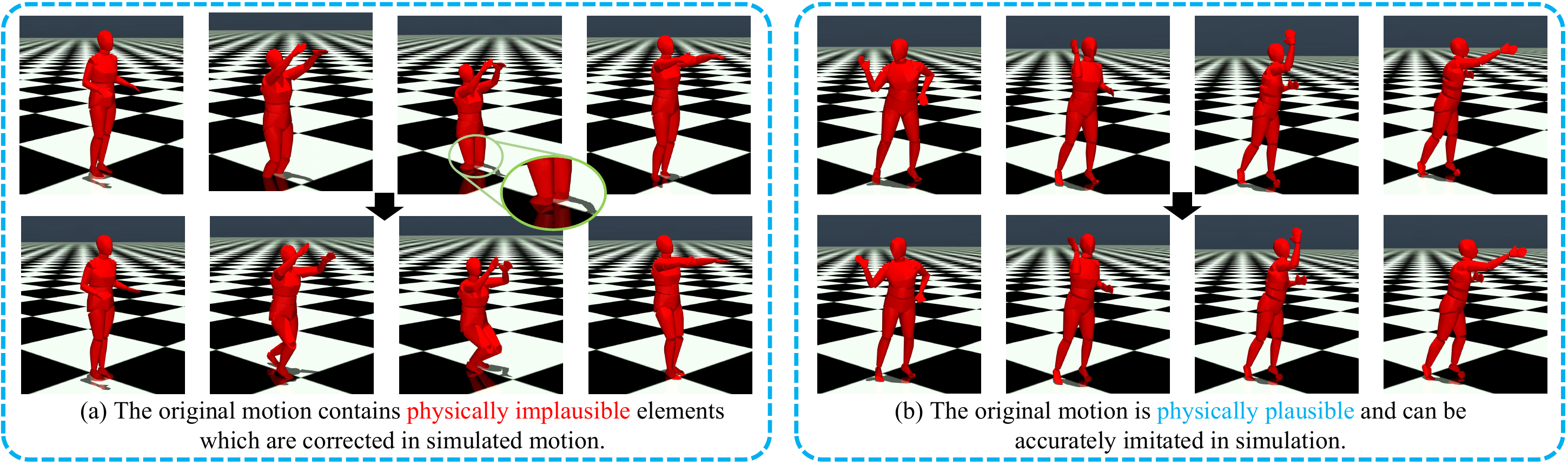}
  \caption{Visualized examples for our dataset labeling. Above: Original motion sequence from the MotionPercept dataset. Below: The motion sequence imitated in the simulator using a per-sequence fine-tuned model.
}
  \vspace{-3mm}
  \label{fig:dataset}
  \Description{Visualized examples for our dataset labeling.}
\end{figure*}

\section{Dataset}
\label{sec:dataset}

% As mentioned in \cref{sec:metric}, we propose a novel annotation method for a fine-grained, interpretable measurement of motion's physical accuracy with the help of a physics simulator. To facilitate our method with the ability for measuring both physical feasibility and human perception, we conduct this physical annotating process on MotionPercept~\cite{motioncritic2025}, which is a large scale dataset of motion perceptual evaluation. Intuitively, the researchers of MotionPercept have invited real humans to select the best or the worst from given motion sets. Each motion sets have 4 motions generated by the state-of-the-art motion generation models MDM~\cite{tevet2022human} and FLAME~\cite{kim2023flame} with the same action label or text prompt. Specifically, the MDM model is trained on HumanAct12~\cite{guo2020action2motion} and UESTC~\cite{ji2018large}, resulting in a total of \textcolor{red}{xx} groups of motions, each containing 4 motions. The FLAME model is trained on HumanML3D~\cite{guo2022generating}, resulting in \textcolor{red}{xx} groups of motions, each group comprising 4 motions. 
% tianyu：seperated to 4.1 and 4.2

\subsection{MotionPercept Dataset}
We use the MotionPercept~\cite{motioncritic2025} dataset to provide perceptual annotations and also provide generated motions for our physical annotation.
MotionPercept is a large-scale dataset of motion perceptual evaluation, in which real humans are invited to select the best or the worst from given motion sets. Each motion set has 4 motions generated by the state-of-the-art motion generation models MDM~\cite{tevet2022human} and FLAME~\cite{kim2023flame} with the same action label or text prompt. Specifically, the MDM model is trained on HumanAct12~\cite{guo2020action2motion} and UESTC~\cite{ji2018large}, resulting in a total of 17521 groups of motions, each containing 4 motions. The FLAME model is trained on HumanML3D~\cite{guo2022generating}, resulting in 201 groups of motions, each group also comprising 4 motions.

\begin{table*}[ht]
\centering
    \caption{Quantitative comparison of our metric with previous metrics. We report human perceptual accuracy \cite{motioncritic2025} and physical correlation PLCC \cite{pearson1920plcc}, SROCC \cite{spearman1910srocc}, and KROCC \cite{loshchilov2017decoupled} on 2 datasets, MotionPercept-MDM and MotionPercept-FLAME. \textbf{Bold} numbers indicate the best results.}
    \label{tab:sota}
    \vspace{-2mm}
    \resizebox{0.8\linewidth}{!}{%
    \begin{tabular}{l|cccc|cccc}
    \hline
    \multirow{2}{*}{Metrics} & \multicolumn{4}{c|}{MotionPercept-MDM} & \multicolumn{4}{c}{MotionPercept-FLAME} \\
    \cline{2-9}
     & Accuracy(\%)$\uparrow$& PLCC$\uparrow$& SROCC$\uparrow$& KROCC$\uparrow$& Accuracy(\%)$\uparrow$& PLCC$\uparrow$& SROCC$\uparrow$& KROCC$\uparrow$\\
     \hline
    Root AVE \cite{motioncritic2025}& 59.47 & 0.323 & 0.223 & 0.150 & 48.43 & 0.048 & 0.135 & 0.089 \\
    Root AE \cite{motioncritic2025}& 61.79 & 0.436 & 0.412 & 0.295 & 59.54 & 0.135 & 0.304 & 0.208 \\
    Joint AVE \cite{motioncritic2025}& 56.77 & 0.322 & 0.239 & 0.164 & 44.61 & 0.072 & 0.112 & 0.079 \\
    Joint AE \cite{motioncritic2025}& 62.73 & 0.467 & 0.456 & 0.327 & 58.37 & 0.236 & 0.377 & 0.262 \\
    PFC \cite{motioncritic2025}& 64.79 & 0.441 & 0.504 & 0.364 & 66.00 & 0.298 & 0.451 & 0.325 \\
    Penetration \cite{ugrinovic2024multiphys}& 50.88 & 0.169 & 0.082 & 0.058 & 56.72 & 0.229 & 0.215 & 0.152 \\
    Skating \cite{ugrinovic2024multiphys}& 52.46 & 0.219 & 0.132 & 0.096 & 56.72 & 0.092 & 0.190 & 0.137 \\
    Floating \cite{ugrinovic2024multiphys}& 55.13 & 0.382 & 0.318 & 0.230 & 55.06 & 0.478 & 0.426 & 0.305 \\
    MotionCritic \cite{motioncritic2025}& 85.07 & 0.329 & 0.316 & 0.220 & 67.66 & 0.152 & 0.280 & 0.188 \\
    \hline
    \model{} (Ours) & \textbf{85.18} & \textbf{0.727} & \textbf{0.622} & \textbf{0.461} & \textbf{68.82} & \textbf{0.657} & \textbf{0.660} & \textbf{0.487}\\
    \hline
    \end{tabular}
    }
\end{table*}

\begin{table*}[ht]
\centering
    \caption{Pearson's Correlation Coefficients (PLCC) results on 12 different prompts on HumanAct12 and on the total 40 prompts of UESTC. HumanAct12 and UESTC are 2 subsets of MotionPercept-MDM. \textbf{Bold} numbers indicate the best results, and {\ul underline} numbers indicate the second best results.}
    \label{tab:sota_plcc}
    \vspace{-2mm}
    \resizebox{0.95\linewidth}{!}{%
    \begin{tabular}{l|cccccccccccc|c|c}
    \hline
    \multirow{2}{*}{Metrics} &  \multicolumn{12}{c|}{HumanAct12}& \multirow{2}{*}{UESTC}& \multirow{2}{*}{Total} \\
    \cline{2-13}
    &  P00&P01& P02& P03& P04& P05& P06& P07 & P08 & P09 & P10 & P11 & &\\
    \hline
    Root AVE \cite{motioncritic2025}&  0.610&0.158 & 0.129 & 0.187 & 0.559 & 0.559 & -0.140 & 0.050 & -0.007 & -0.154 & 0.458 & 0.181 & 0.355 & 0.323 \\
    Joint AVE \cite{motioncritic2025}&  0.509&0.115 & 0.177 & 0.344 & 0.456 & 0.576 & -0.127 & \textbf{0.278} & -0.004 & -0.141 & 0.571 & 0.018 & 0.350 & 0.322 \\
    Joint AE \cite{motioncritic2025}&  {\ul0.738}&0.455 & {\ul 0.555} & 0.220 & 0.535 & 0.552 & -0.182 & -0.010 & -0.047 & -0.229 & {\ul 0.647} & 0.147 & 0.522 & {\ul 0.467} \\
    Root AE \cite{motioncritic2025}&  0.714&{\ul 0.564} & 0.475 & 0.171 & {\ul 0.568} & {\ul 0.642} & -0.242 & 0.181 & -0.019 & -0.240 & 0.629 & 0.184 & 0.476 & 0.436 \\
    PFC \cite{motioncritic2025}&  0.521&0.099 & 0.330 & 0.286 & 0.478 & 0.587 & 0.458 & -0.044 & -0.012 & 0.100 & 0.437 & 0.255 & {\ul 0.486} & 0.441 \\
    Penetration \cite{ugrinovic2024multiphys}&  0.220&-0.190 & -0.137 & 0.005 & 0.155 & 0.275 & -0.394 & 0.173 & -0.116 & 0.059 & -0.120 & 0.219 & 0.216 & 0.169 \\
    Skating \cite{ugrinovic2024multiphys}&  0.320&-0.257 & -0.045 & -0.039 & 0.371 & 0.201 & -0.226 & -0.090 & -0.077 & -0.171 & 0.078 & 0.271 & 0.277 & 0.219 \\
    Floating \cite{ugrinovic2024multiphys}&  0.601&0.341 & 0.534 & \textbf{0.594} & 0.344 & 0.570 & {\ul 0.754} & 0.021 & {\ul -0.003} & -0.004 & \textbf{0.664} & {\ul 0.425} & 0.375 & 0.382 \\
    MotionCritic \cite{motioncritic2025}&  0.385&0.525 & 0.438 & 0.096 & 0.328 & 0.334 & 0.688 & -0.284 & -0.274 & {\ul 0.163} & 0.223 & 0.217 & 0.302 & 0.287 \\
    \hline
    \model{} (Ours) &  \textbf{0.760}&\textbf{0.983} & \textbf{0.808} & {\ul 0.541} & \textbf{0.699} & \textbf{0.664} & \textbf{0.782} & {\ul 0.272} & \textbf{0.123} & \textbf{0.663} & 0.568 & \textbf{0.515} & \textbf{0.760} & \textbf{0.727} \\
    \hline
    \end{tabular}
    }
\end{table*}

\begin{table*}[ht]
\centering
    % \caption{We report the separated Spearman's Ranking Order Correlation Coefficients (SROCC) results on 12 different prompts on HumanAct12 and on the total 40 prompts of UESTC.}
    \caption{Spearman's Ranking Order Correlation Coefficients (SROCC) on 12 different prompts on HumanAct12 and on UESTC.}
    \label{tab:sota_srocc}
    \vspace{-2mm}
    \resizebox{0.95\linewidth}{!}{%
    \begin{tabular}{l|cccccccccccc|c|c}
    \hline
    \multirow{2}{*}{Metrics} & \multicolumn{12}{c|}{HumanAct12}& \multirow{2}{*}{UESTC}& \multirow{2}{*}{Total} \\
    \cline{2-13}
    & P00&P01 & P02 & P03 & P04 & P05 & P06 & P07 & P08 & P09 & P10 & P11 & &\\
    \hline
    Root AVE \cite{motioncritic2025}& 0.450&0.052 & -0.185 & 0.037 & 0.372 & 0.418 & -0.280 & 0.081 & {\ul 0.104} & -0.292 & 0.166 & 0.262 & 0.260 & 0.223 \\
    Joint AVE \cite{motioncritic2025}& 0.085&0.011 & -0.161 & 0.235 & 0.040 & 0.304 & -0.073 & \textbf{0.357} & -0.117 & -0.316 & 0.338 & 0.100 & 0.290 & 0.239 \\
    Joint AE \cite{motioncritic2025}& 0.612&0.591 & 0.578 & 0.292 & 0.301 & 0.485 & 0.059 & -0.028 & -0.207 & -0.310 & 0.339 & 0.203 & 0.520 & 0.456 \\
    Root AE \cite{motioncritic2025}& {\ul0.632}&0.537 & 0.561 & 0.304 & 0.509 & 0.469 & -0.055 & {\ul 0.183} & -0.184 & -0.376 & 0.269 & 0.294 & 0.458 & 0.412 \\
    PFC \cite{motioncritic2025}& \textbf{0.659}&0.404 & {\ul 0.632} & 0.350 & {\ul 0.533} & {\ul \textbf{0.610}} & 0.601 & 0.004 & 0.015 & -0.197 & \textbf{0.640} & {\ul 0.330} & {\ul 0.541} & {\ul 0.504} \\
    Penetration  \cite{ugrinovic2024multiphys}& 0.090&-0.394 & -0.069 & -0.152 & 0.199 & 0.032 & -0.285 & 0.127 & -0.172 & -0.137 & -0.048 & 0.121 & 0.124 & 0.082 \\
    Skating \cite{ugrinovic2024multiphys}& 0.156&-0.390 & 0.018 & -0.145 & 0.317 & -0.007 & -0.057 & -0.087 & -0.022 & -0.276 & 0.194 & 0.227 & 0.173 & 0.132 \\
    Floating \cite{ugrinovic2024multiphys}& 0.330&\textbf{0.760} & 0.467 & \textbf{0.569} & 0.176 & {\ul 0.528} & {\ul 0.716} & -0.020 & \textbf{0.109} & 0.006 & 0.274 & 0.236 & 0.310 & 0.318 \\
    MotionCritic \cite{motioncritic2025}& 0.450&0.460 & 0.400 & 0.071 & 0.402 & 0.373 & 0.696 & -0.357 & -0.323 & {\ul 0.052} & 0.288 & 0.249 & 0.299 & 0.283 \\
    \hline
    \model{} (Ours) & 0.551&{\ul 0.593} & \textbf{0.716} & {\ul 0.435} & \textbf{0.596} & 0.455 & \textbf{0.778} & -0.156 & -0.013 & \textbf{0.203} & {\ul 0.426} & \textbf{0.545} & \textbf{0.681} & \textbf{0.622} \\
    \hline
    \end{tabular}
    }
\end{table*}

\subsection{Physical Annotation Generation}
% \textcolor{red}{ToDO}

% We propose a novel annotation method for a fine-grained, interpretable measurement of motion's physical accuracy with the help of a physics simulator. To facilitate our method with the ability for measuring both physical feasibility and human perception, we conduct this physical annotating process on MotionPercept~\cite{motioncritic2025}

% We utilize a series of metrics to quantify the discrepancy between the motions imitated in Isaac simulator and the ground-truth motions in \textit{MotionPercept} dataset. Reconstruction error(Recon. Err.) calculates the absolute mean per-joint position error in world coordinates. MPJPE (Mean Per Joint Position Error) measures the mean per-joint position error relative to the root joint, while PA-MPJPE (Procrustes-Aligned MPJPE) evaluates position error after optimal rigid alignment. We also compute acceleration error $e_{\text{acc}}$ and velocity error $e_{\text{vel}}$.

We propose a novel annotation method that leverages a physics simulator to provide a fine-grained and interpretable measurement of a motion's physical accuracy. Specifically, we use the simulator to identify a physically plausible motion $x^\prime$ that is as close as possible to the input motion $x$. The physical error $e_p$ is then defined as the $l_2$ norm of the difference between the two motions: 
\begin{equation}
    e_p = \|x-x^\prime\|_2,
\end{equation}
where $\|\cdot\|_2$ is $l_2$ norm. To obtain an $x^\prime$ that closely approximates $x$, we propose using a physical correction network, $F_p(x)$, is defined as:
\begin{equation}
    x^\prime = F_p(x),
\end{equation}
where $x^\prime$ is the output motion that is aligned with physical laws in physical simulators. 

The training of $F_p(x)$ would require feedback from a physical simulator. Since most state-of-the-art simulators are not designed to backpropagate gradients, we used reinforcement learning to get the physical fidelity feedback from the physical simulator. Moreover, we also constrain the output motion $x^\prime$ to be close to the input motion $x$ in terms of translation, rotation, linear velocity, and angular velocity.
Specifically, we use the network in PHC \cite{Luo2023PerpetualHC} as our physical correction network to find $x^\prime$ for each pose in the MotionPercept~\cite{motioncritic2025} dataset. To achieve a lower difference while still aligning with physical constraints, we use the physical reward in PHC \cite{Luo2023PerpetualHC}. For timestamp $t$, the reward can be represented as:
\begin{equation}
\begin{array}{cc}
    &r_t= w_{jp}e^{-100\|p^\prime_t-p_t\|}
    + w_{jr}e^{-10\|q^\prime_t \ominus q_t\|} \\
    &+ w_{jv}e^{-0.1\|v^\prime_t-v_t\|}
    + w_{j\omega}e^{-0.1\|\omega^\prime_t-\omega_t\|},
\end{array}
\label{eq:phc_loss}
\end{equation}
where $p_t$, $q_t$, $v_t$, and $\omega_t$ are the translation, rotation, linear velocity, and angular velocity of input motion $x$ at timestamp $t$. $p^\prime_t$, $q^\prime_t$, $v^\prime_t$, and $\omega^\prime_t$ are the translation, rotation, linear velocity, and angular velocity of the output motion $x^\prime$ from the physical simulator at timestamp $w_{jp}$. $w_{jr}$, $w_{jv}$, and $w_{j\omega}$ are the loss weights. $\ominus$ is the difference between 2 rotations. This reward function makes sure the physical simulator output is close to the original pose, and still aligns with physical laws. In training, we summarize the reward function for all timestamps $t$ and get the optimal motion $x^\prime$.

Our annotation generation contains 2 steps: First, we pretrain the PHC network on the whole MotionCritic dataset using the reward function \cref{eq:phc_loss}. Second, we use the same reward function to optimize every single motion in the MotionCritic dataset, so that it will achieve a closer corrected motion for each input motion in the dataset. In \cref{tab:ft_mdm} we report the difference between the input motion $x$ and the corrected motion $x^\prime$ after the whole dataset pretrain (step 1) and the per-data fine-tune (step 2), respectively. In these tables, we use IsaacGym \cite{makoviychuk2021isaac} simulator for physical simulation. Reconstruction error (Recon. Err.) calculates the absolute mean per-joint position error in world coordinates. MPJPE (Mean Per Joint Position Error) measures the mean per-joint position error relative to the root joint, while PA-MPJPE (Procrustes-Aligned MPJPE) evaluates position error after optimal rigid alignment. We also compute acceleration error $e_{\text{acc}}$ and velocity error $e_{\text{vel}}$.
From the results, we observe that our per-data fine-tune result provides closer corrected motion than the whole data pretrain step.

In \cref{tab:dataset}, we also provide simple statistics and a comparison with the previous annotation in MotionPercept. Our annotation is fine-grained, physically aligned, and is normalized to a $N(0,1)$ normal distribution. In \cref{fig:dataset}, we visualize the input motion along with the final fine-tuned corrected motion. Our corrected motion well aligns with physical laws.

\begin{table*}[ht]
\centering
    % \caption{We report the separated Kendall's Ranking Order Correlation Coefficients (KROCC) results on 12 different prompts on HumanAct12 and on the total 40 prompts of UESTC. }
    \caption{Kendall's Ranking Order Correlation Coefficients (KROCC) on 12 different prompts on HumanAct12 and on UESTC.}
    \label{tab:sota_krocc}
    \vspace{-2mm}
    \resizebox{0.95\linewidth}{!}{%
    \begin{tabular}{l|cccccccccccc|c|c}
    \hline
    \multirow{2}{*}{Metrics} &  \multicolumn{12}{c|}{HumanAct12}& \multirow{2}{*}{UESTC}& \multirow{2}{*}{Total} \\
    \cline{2-13}
    & P00 & P01 & P02 & P03 & P04 & P05 & P06 & P07 & P08 & P09 & P10 & P11 &  &\\
    \hline
    Root AVE \cite{motioncritic2025}&  0.316&0.043 & -0.139 & 0.032 & 0.249 & 0.286 & -0.189 & 0.065 & \textbf{0.075} & -0.202 & 0.102 & 0.166 & 0.175 & 0.150 \\
    Joint AVE \cite{motioncritic2025}&  0.063&0.014 & -0.112 & 0.157 & 0.028 & 0.202 & -0.068 & \textbf{0.247} & -0.058 & -0.222 & 0.246 & 0.057 & 0.199 & 0.164 \\
    Joint AE \cite{motioncritic2025}&  0.433&0.425 & 0.419 & 0.213 & 0.206 & {\ul 0.339} & 0.048 & -0.021 & -0.146 & -0.186 & 0.236 & 0.147 & 0.373 & 0.327 \\
    Root AE \cite{motioncritic2025}&  {\ul0.457}&0.387 & 0.405 & 0.217 & 0.355 & 0.326 & -0.033 & {\ul 0.116} & -0.126 & -0.244 & 0.199 & 0.202 & 0.326 & 0.295 \\
    PFC \cite{motioncritic2025}&  \textbf{0.482}&0.275 & {\ul 0.454} & 0.230 & {\ul 0.386} & 0.420 & 0.405 & 0.016 & 0.007 & -0.128 & \textbf{0.438} & {\ul 0.230} & {\ul 0.392} & {\ul 0.364} \\
    Penetration \cite{ugrinovic2024multiphys}&  0.057&-0.271 & -0.049 & -0.111 & 0.136 & 0.021 & -0.203 & 0.084 & -0.128 & -0.095 & -0.027 & 0.086 & 0.088 & 0.058 \\
    Skate \cite{ugrinovic2024multiphys}&  0.107&-0.272 & 0.015 & -0.101 & 0.225 & -0.007 & -0.025 & -0.077 & -0.008 & -0.200 & 0.133 & 0.165 & 0.125 & 0.096 \\
    Float \cite{ugrinovic2024multiphys}&  0.236&\textbf{0.572} & 0.308 & \textbf{0.409} & 0.123 & \textbf{0.373} & {\ul 0.521} & -0.017 & {\ul 0.072} & 0.009 & 0.196 & 0.161 & 0.225 & 0.230 \\
    MotionCritic \cite{motioncritic2025}&  0.318&0.315 & 0.278 & 0.038 & 0.279 & 0.245 & 0.507 & -0.232 & -0.221 & {\ul 0.038} & 0.191 & 0.167 & 0.208 & 0.197 \\
    \hline
    \model{} (Ours) &  0.398&{\ul 0.431} & \textbf{0.515} & {\ul 0.300} & \textbf{0.435} & 0.320 & \textbf{0.586} & -0.101 & -0.015 & \textbf{0.140} & {\ul 0.285} & \textbf{0.349} & \textbf{0.509} & \textbf{0.461} \\
    \hline
    \end{tabular}
    }
\end{table*}

\begin{table*}[ht]
\centering
    \caption{Ablation studies on different loss functions and training strategies.}
    \label{tab:abla}
    \vspace{-2mm}
    \resizebox{0.85\linewidth}{!}{%
    \begin{tabular}{l|cccc|cccc}
    \hline
    \multirow{2}{*}{Metrics}  & \multicolumn{4}{c|}{MotionPercept-MDM} & \multicolumn{4}{c}{MotionPercept-FLAME} \\
     \cline{2-9}
    \multirow{-2}{*}{\textbf{}} & Accuracy(\%)$\uparrow$& SROCC$\uparrow$& KROCC$\uparrow$& PLCC$\uparrow$& Accuracy(\%)$\uparrow$& SROCC$\uparrow$& KROCC$\uparrow$& PLCC$\uparrow$\\
    \hline
    MotionCritic & 85.07 & 0.3290 & 0.3160 & 0.2200 & 67.66 & 0.1520 & 0.2797 &0.1875 \\
    \hline
    w/o prompt categorization & \textbf{85.61} & 0.5191 & 0.3791 & 0.6146 & \textbf{70.98} & 0.6422 & 0.4649 & 0.6347 \\
    MSE loss & 84.29 & 0.6000& 0.4446 & 0.6357 & 69.48 & 0.6059 & 0.4446 & 0.5797 \\
    \hline
    \model{} (Ours) & 85.18 & \textbf{0.6223} & \textbf{0.4612} & \textbf{0.7268} & 68.82 & \textbf{0.6598} & \textbf{0.4873} & \textbf{0.6567} \\
    \hline
    \end{tabular}
    }
\end{table*}

\begin{table}[ht]
\centering
    \caption{Comparison of MDM model performance before and after finetuning with our metric. PP-Motion is the average predicted score of our metric. Mean MPJPE is the mean per-joint position error between simulated motion and ground truth motion.} 
    \label{tab:exp_ft}
    % \vspace{-2mm}
    \resizebox{0.85\linewidth}{!}{%
    \begin{tabular}{l|cc}
    \hline
     & PP-Motion$\uparrow$ & Mean MPJPE$\downarrow$\\
     \hline
    Before Fine-tuning & -0.09 & 76.06 \\
    Fine-tune 100 steps & 0.61 & 63.33 \\
    \hline
    \end{tabular}
    }
\end{table}

\begin{figure*}[ht]
  \centering
  % \vspace{-5mm}
  \includegraphics[width=0.93\linewidth]{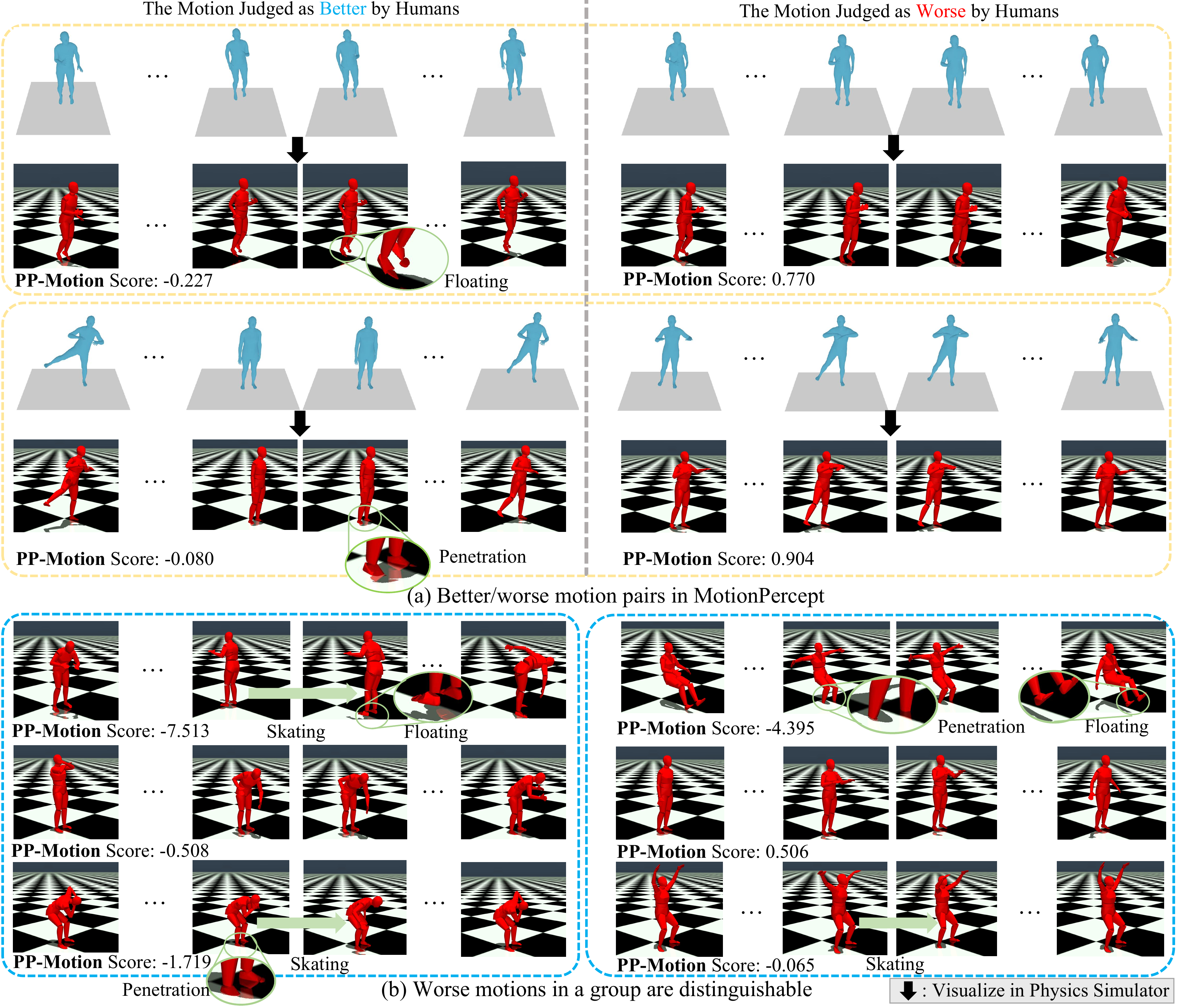}
  % \vspace{1mm}
  \caption{(a) 
A MotionPercept better/worse data pair: the motion on the left, annotated as `better' and visually superior, exhibits physics issues (e.g. floating, skating). The motion on the right, annotated as `worse' and visually inferior, shows greater physical plausibility.
(b) 
Three MotionPercept samples in a group, all annotated as `worse' by humans, reveal different physical characteristics in the simulator. Our \model{} scores (higher means better) successfully capture these physical distinctions. 
}
  
  \label{fig:vis}
  \Description{Visualization}
\end{figure*}

\section{Experiment Results}

% \subsection{Comparison with MotionCritic}
\textbf{Comparison with previous metrics.}
We verify the human perceptual and physical alignment of our metric PP-Motion and previous metrics on 2 subsets of MotionPercept: MDM and FLAME. The results are shown in \cref{tab:sota}. For human perceptual alignment, we evaluated the ``better/worse'' classification accuracy. For physical alignment, we evaluated 3 correlation methods (i.e., PLCC~\cite{pearson1920plcc}, SROCC~\cite{spearman1910srocc}, KROCC~\cite{loshchilov2017decoupled} between \model{} results and physical annotation. The correlation methods are defined in supplementary materials. Our metric is trained only on the MDM training set and directly tested on the MotionPercept-MDM validation and the MotionPercept-FLAME dataset without further finetuning. The results show that our metric outperforms previous works on physical alignment. Notably, on human perceptual alignment, our metric slightly outperforms the baseline method, MotionCritic, which proves that the physical annotation has the potential to improve the human perceptual alignment of our metric. We also compare our metric with existing pose-based metrics (i.e., Root AVE, Root AE, Joint AVE, Joint AE) and physics-based metrics (i.e., Penetration, Float, Skate).  Detailed definitions of these metrics are provided in supplementary materials.
% The correlation methods are defined in \cref{eq:plcc2}, \cref{eq:srocc}, and \cref{eq:krocc}. 

\textbf{Per-category physical alignment.} As shown in \cref{tab:sota_plcc}, \cref{tab:sota_srocc}, and \cref{tab:sota_krocc}, we further report the per-category correlation results on 12 different prompts of HumanAct12 \cite{guo2020action2motion} and 40 prompts of UESTC \cite{ji2018large}. We observe that on most categories, our metric has the best or the second best correlation with the physical annotation among all metrics. This further proves the generalizability of our metric.

\textbf{Ablation studies.}
We perform ablation studies to validate our metric designs, with results presented in \cref{tab:abla}. First, in ``\texttt{w/o prompt categorization}'', we examine the impact of a prompt-categorized training strategy by training directly on the MotionPercept-MDM dataset without categorizing by prompt labels. The prompt-categorized training approach is detailed in supplementary materials. The results demonstrate that computing PLCC loss within the same-label motion groups achieves better results in physical correlation metrics. Second, in ``\texttt{MSE loss}'', we train the metric with MSE loss, which calculates the $l_2$ distance between the predicted score and GT annotation. The results show that replacing PLCC loss with conventional MSE optimization leads to noticeable degradation in both physical plausibility assessment and human evaluation metrics. 

\textbf{Improving motion generation with \model{}.}
We try to improve the motion generation method MDM \cite{tevet2022human} to further verify our metric's physical fidelity alignment. We fine-tune the MDM network using our metric, along with critic loss and KL loss from \cite{motioncritic2025}. The critic loss for the input pose $x'_0$ is defined as:
\begin{equation}
\mathcal{L}_{\text{Critic}} = \mathbb{E}\left[-\sigma\left(\tau-F\left(x'_0\right)\right)\right],
\end{equation}
where $\tau$ is the threshold of sigmoid function $\sigma(\cdot)$, and $\mathbb{E}$ is the expectation on the whole dataset. The KL loss is defined as
\begin{equation}
\mathcal{L}_{\text{KL}} = \mathbb{E}\left[D_{\text{KL}}\Bigl(p(x'_0) \,|\, p(\tilde{x'_0})\Bigr)\right],
\end{equation}
where $D_{\text{KL}}$ is KL divergence, and $\tilde{x'_0}$ is the pervious iteration of $x'_0$.

We fine-tune the MDM model for 100 steps and generate 120 motion sequences each using the MDM baseline model and the fine-tuned model. Then we fine-tune each motion on PHC, following the procedure described in \cref{sec:dataset}, and calculate the mean MPJPE between the motion simulated in Isaac and the ground truth motion generated by MDM. The results reported in \cref{tab:exp_ft} show that our metric can improve physical alignment in motion generation.

\textbf{Visualization.}
% \cref{fig:vis} \textcolor{red}{TODO}
\cref{fig:vis} (a) shows a better/worse data pair sampled from the MotionPercept dataset. 
The motion on the left (annotated as `better' and visually superior) exhibits physics issues (e.g. floating, skating) in the simulator, while the motion on the right (annotated as `worse' and visually inferior) demonstrates greater physical plausibility when simulated.
\cref{fig:vis} (b) shows three MotionPercept samples in a group that are all annotated as `worse', but reveal different physical characteristics in the simulator. Our \model{} scores successfully capture these physical distinctions. Note that for our PP-Motion score, the higher reveals the better.

\section{Conclusion}
In this work, we address the challenges in evaluating the fidelity of generated human motions by bridging the gap between human perception and physical feasibility. We introduce a novel physical labeling method that computes the minimum adjustments needed for a motion to adhere to physical laws, thereby producing fine-grained, continuous physical alignment annotations as objective ground truth. Our framework leverages Pearson's correlation loss to capture the underlying physical priors, while integrating a human-based perceptual fidelity loss to ensure that the evaluation metric reflects both physical fidelity and human perception. Experimental results validate that our metric not only complies with physical laws but also demonstrates superior alignment with human perception compared to previous approaches.

\clearpage

\begin{acks}
This work is supported by the National Key R\&D Program of China under Grant No.2024QY1400, and the National Natural Science Foundation of China No. 62425604. This work is also supported by Tsinghua University Initiative Scientific Research Program and the Institute for Guo Qiang at Tsinghua University.
\end{acks}

\bibliographystyle{ACM-Reference-Format}
\bibliography{main}

\appendix

\section{Implementation Details}
We train our \model{} model using the MDM subset of MotionPercept, which contains 46761 better-worse motion pairs. The MDM dataset is generated from 12 action labels in HumanAct12 and 40 labels in UESTC. We first categorize the dataset according to 52 prompt labels. During training, each batch contains motions exclusively from the same category, and the PLCC loss is computed collectively across all motions within a batch. Our model architecture follows \cite{motioncritic2025}, employing a DSTformer \cite{zhu2023motionbert} backbone with 3 layers and 8 attention heads for generating motion embedding. The embeddings are then fed into a 1024-channel MLP layer that produces a single scalar score. We train the \model{} model for 200 epochs with a batch size of 64 and a learning rate initialized at 4e-5 and decayed exponentially (factor 0.995 per epoch). The correlation loss term is weighted by $\lambda=0.3$.

\section{Evaluation Methods}
To evaluate how well our metric corresponds with human perception, we leverage three distinct correlation measures. First, Pearson’s linear correlation coefficient (PLCC) \cite{pearson1920plcc} quantifies the linear association between our metric's predicted scores and physical annotations. It is calculated as :
\begin{equation}
C_p=\frac{\sum_{i=1}^n(\hat{x}_i-\bar{\hat{x}})(x_i-\bar{x})}{\sqrt{\sum_{i=1}^n(\hat{x}_i-\bar{\hat{x}})^2} \sqrt{\sum_{i=1}^n(x_i-\bar{x})^2}},
\label{eq:plcc2}
\end{equation}
where $\hat{x}_i$ and $x_i$ indicate the predicted and ground truth motion fidelity scores of the sample $i$ in the dataset. $n$ is the total number of samples in the dataset. The average predicted motion fidelity and ground truth motion fidelity are then defined as $\bar{\hat{s}}=\frac{1}{n}\sum_{i=1}^n\hat{s}_i$ and $\bar{s}=\frac{1}{n}\sum_{i=1}^n s_i$.

Next, we employ Spearman's ranking order correlation coefficient (SROCC) \cite{spearman1910srocc} to assess the agreement in ranking between our metric and physical annotations. SROCC is expressed as: 
\begin{equation} 
C_s=1-\frac{6\sum_{i=1}^n(R(\hat{x}_i)-R(x_i))^2}{n(n^2-1)}, 
\label{eq:srocc}
\end{equation}
where \( R(\hat{x}_i) \) and \( R(x_i) \) indicate the ranks of $\hat{x}_i$ and $x_i$, and $n$ is the number of data points.

Finally, we use Kendall’s rank order correlation coefficient (KROCC) \cite{loshchilov2017decoupled} to further verify the ranking consistency between our metric and physical perception. Different from SROCC, Kendall’s coefficient focuses only on the concordance of rank order. It is given by: 
\begin{equation}
C_\tau = 1 - \frac{2}{n(n^2 - 1)} \sum_{i<j} \operatorname{sign}(\hat{x}_i - \hat{x}_j)\operatorname{sign}(x_i - x_j).
\label{eq:krocc}
\end{equation}
The sign function $\operatorname{sign}(\cdot)$ is defined as:
\begin{equation}
\operatorname{sign}(x) =
\begin{cases} 
\frac{x}{|x|} & x \neq 0 \\ 
0, & x = 0 
\end{cases}
\end{equation}

All three coefficients range from -1 to 1, with values closer to 1 indicating a stronger correlation.

\section{Previous metrics}
Existing metrics to evaluate the quality of a motion sequence can be categorized into two main approaches: (1) error-based metrics, which quantify the discrepancy between a pair of generated motion and ground-truth (GT) motion, and (2) physics-based metrics, which assess the physical plausibility of a motion sequence. For error-based metrics, we report Root Average Variance Error (Root AVE), Root Absolute Error (Root AE), Joint Average Variance Error (Joint AVE), and Joint Absolute Error (Joint AE) following \cite{motioncritic2025}. Absolute Error (AE) computes the average L2 distance between corresponding joint positions in generated and ground truth motions. Average Variance Error (AVE) measures the average L2 distance between the per-joint temporal variance of generated motions and that of the ground truth motions, reflecting how well the dynamics of the motion are captured over time. For physics-based metrics, we report ground penetration (Penetration), foot skating (Skate), and floating (Float) following \cite{ugrinovic2024multiphys}. We also report Physical Foot Contact (PFC) following \cite{motioncritic2025}.

\section{Visualization}

\subsection{Dataset visualization}

% Videos corresponding to Figure 3 in the main text can be found in Vis\_dataset folder, which contains two subfolders, figure3(a) and figure3(b). 
We visualize both the raw motion data from MotionPercept and the imitated motion sequences using per-data finetuned model in physics simulator, which are shown in \cref{fig:dataset}. The results demonstrate that when the original motion contains physically implausible elements, the imitation process corrects these artifacts. Conversely, for physically valid raw motions, the imitated sequences accurately preserve the kinematic characteristics of the raw motions.

\subsection{Visualize cases}

% The video demonstrations corresponding to Figure 4 in the main text are available in the Vis\_cases folder, which comtains two subfolders, figure4(a) and figure4(b) . 

We provide visualized cases from the MotionPercept dataset in \cref{fig:vis} to verify the effectiveness of our \model{} metric.

\cref{fig:vis} (a) contains two examples, each visualizing a better/worse data pair from the MotionPercept dataset. When rendered in 2D visualization videos, the better cases appear visually superior to human observers. However, simulator-based visualization reveals that the better cases actually contain more physically implausible artifacts:
\begin{itemize}
    \item Example 1: `better' motion exhibits floating
    % \item Example 2: `better' motion shows skating artifacts
    \item Example 2: `better' motion demonstrates both skating and penetration phenomena
\end{itemize}
Our \model{} scoring effectively captures these physical plausibility considerations:
\begin{itemize}
    \item Example 1: `better' motion scored -0.227, `worse' motion scored 0.770
    % \item Example 2: `better' motion scored -0.003, `worse' motion scored 0.568
    \item Example 2: `better' motion scored -0.080, `worse' motion scored 0.904
\end{itemize}

\cref{fig:vis} (b) contains two examples, each visualizing three motions that come from the same annotation group in MotionPercept. Human visual assessment found these motions to be of similar quality, and they are all annotated as `worse' samples in MotionPercept dataset. However, their adherence to physical laws varies significantly, as observed in physics simulation.
\begin{itemize}
    \item Example 1: floating and skating in ``worse1'' motion, penetration and skating in ``worse3'' motion.
    \item Example 2: floating and penetration in ``worse1'' motion, skating in ``worse3'' motion.
\end{itemize}

\model{} scoring demonstrates strong correlation with physical feasibility assessments:
\begin{itemize}
    \item  Example 1: scores for ``worse1'', ``worse2'', and ``worse3'' are -7.513, -0.508 and -1.719.
    \item Example 2: scores for ``worse1'', ``worse2'', and ``worse3'' are -4.395, 0.506, and -0.065.
\end{itemize}

% \bibliographystyle{ACM-Reference-Format}

% \appendix

% \end{document}

\end{document}